# Zero-shot large vision-language model prompting for automated bone identification in paleoradiology x-ray archives

Owen Dong[*,a], Lily Gao[b], Manish Kota[a], Bennett A. Landman[a,c-f], Jelena Bekvalac[h], Gaynor Western[i], Katherine D. Van Schaik[c,e,g]

[a]Dept. of Computer Science, Vanderbilt University, Nashville, TN, USA; [b]Vanderbilt University School of Medicine, Nashville, TN, USA; [c]Dept. of Electrical and Computer Engineering, Vanderbilt University, Nashville, TN, USA; [d]Vanderbilt University Institute of Imaging Science, Vanderbilt University Medical Center, Nashville, TN, USA; [e]Dept. of Radiology and Radiological Sciences, Vanderbilt University Medical Center, Nashville, TN, USA; [f]Dept. of Biomedical Engineering, Vanderbilt University, Nashville, TN, USA; [g]Dept. of Classical and Mediterranean Studies, Vanderbilt University, Nashville, TN, USA; [h]Museum of London, London, United Kingdom; [i]Ossafreelance, Wisbech, Cambridgeshire, United Kingdom

## ABSTRACT

Paleoradiology, the use of modern imaging technologies to study archaeological and anthropological remains, offers new windows on millennial-scale patterns of human health. Unfortunately, the radiographs collected during field campaigns are heterogeneous: bones are disarticulated, positioning is ad-hoc, and laterality markers are often absent. Additionally, factors such as age-at-death, age-of-bone, sex, and imaging equipment introduce high variability. Thus, content navigation, such as identifying a subset of images with a specific projection view, can be time-consuming and difficult, making efficient triaging a bottleneck for expert analysis. We report a zero-shot prompting strategy that leverages a state-of-the-art Large Vision–Language Model (LVLM) to automatically identify the main bone, projection view, and laterality in such images. Our pipeline converts raw DICOM files to bone-windowed PNGs, submits them to the LVLM with a carefully engineered prompt, and receives structured JSON outputs, which are extracted and formatted onto a spreadsheet in preparation for validation. On a random sample of 100 images reviewed by an expert board-certified paleoradiologist, the system achieved 92% main-bone accuracy, 80% projection-view accuracy, and 100% laterality accuracy, with low- or medium-confidence flags for ambiguous cases. These results suggest that LVLMs can substantially accelerate code-word development for large paleoradiology datasets, allowing for efficient content navigation in future anthropology workflows.

## 1. INTRODUCTION

The radiologic examination of ancient skeletal remains has evolved from rare museum-based fluoroscopy in the early twentieth century to routine digital radiography during archaeological excavations.[1] Portable flat-panel detectors and lightweight X-ray tubes have democratized data collection, but they have also introduced extreme variability.[2,3] Imaging devices differ in exposure setting and operators have different imaging techniques.[4] Additionally, bone morphology changes significantly depending on sex, age-at-death, and age-of-bone, introducing even more variability.[4] Dealing with archaeological remains also increases heterogeneity as bones are often fragmented and field operators seldom follow clinical positioning standards; bones are often imaged in isolation, stacked, or partially obscured by packing materials.[2,4–6] Conventional Picture-Archiving and Communication Systems (PACS) capture basic acquisition metadata, but the content of each image as displayed in Figure 1, such as what bone or skeletal region is present, whether the projection is anteroposterior (AP) or lateral, and which side of the body is represented, remains unavailable unless manually annotated.[3,6] As repositories surpass hundreds of thousands of paleoradiology images, the bottleneck has shifted from acquisition to curation: selecting, organizing, and annotating relevant images for analysis, as the variability present in the images makes this process inefficient and inconsistent.[7]

---

[*] owen.dong@vanderbilt.edu

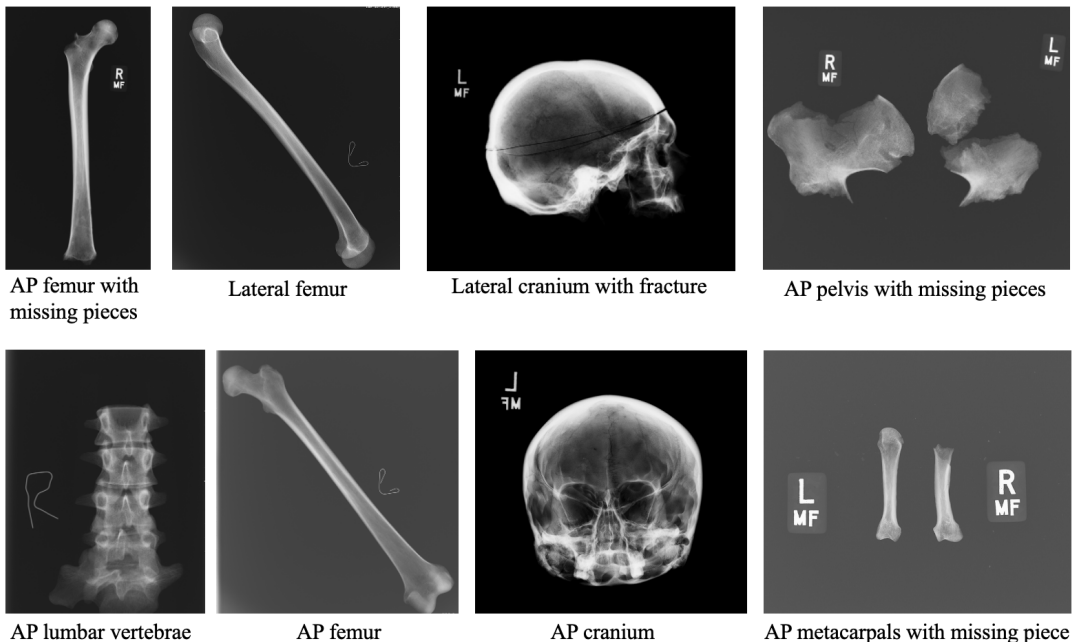

Figure 1. Collection of archaeological radiographs illustrating the diversity of the content encountered in paleoradiology archives. Note: These archaeological images generally lack projection view encoded into the DICOM's metadata, consistent with prior reports that field radiographs of dry bone often omit positioning details.[8]

Automated content labeling of modern clinical radiographs has benefited from convolutional neural networks (CNN) trained on millions of annotated hospital images[9]. However, not only is the annotation process extremely tedious and time-consuming for radiologists, the nature of data between clinical and archaeological data is profound; in latter images, bones are often fragmented, soft-tissue context is absent, and projection geometry is unconstrained, limiting the usefulness of these models if a large, annotated set of images is absent.[4,6,7,10,11] Collecting task-specific training sets with expert labels would re-introduce the overhead we seek to avoid. However, recent large vision–language models (LVLM), trained on diverse internet-scale corpora, exhibit impressive zero-shot reasoning: they can respond to verbose prompts with structured answers even for tasks outside their nominal training distribution.[12] This study hypothesized that such models, when guided by a carefully engineered system prompt, could identify key properties of skeletal structures in paleoradiology images without additional training, eliminating the time requirements needed for expert annotation.[4,7]

Here we describe our end-to-end pipeline, implemented within the Vanderbilt Lab for Immersive AI Translation (VALIANT) multi-platform AI environment. We detail the preprocessing steps required to translate raw DICOM files into LVLM-ready PNGs, the design of the zero-shot prompt, and a validation protocol that leverages an expert paleo-radiologist to score outputs for accuracy. We report preliminary performance metrics, analyze failure modes, and discuss future directions for integrating LVLM outputs into anthropology workflows.

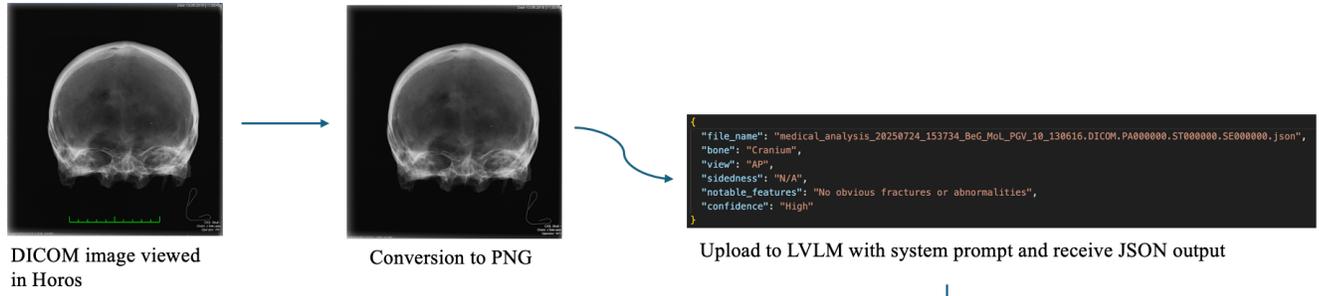

Figure 2. End-to-end workflow. A master folder of raw DICOM images, here viewed in Horos,[13] are first converted into bone-windowed PNGs. The master folder of PNG's acts as our input and each image is queried into the LVLM with our engineered system prompt. The model returns a carefully structured JSON output into a designated output folder. A separate Python script tags each JSON with a numerical tag for sorting purposes and appends the contents to an Excel spreadsheet. This process replaces manual annotation with an automated LVLM system, enabling efficient annotation in large paleoradiology datasets.

## 2. METHODS

### 2.1 Dataset collection
Radiographs were derived from imaging performed on selected skeletonized remains from eleven historical excavation sites around London and were provided by the Museum of London. In total, 8423 DICOM files, were transferred to Vanderbilt's Advanced Computing Center for Research and Education (ACCRE).

### 2.2 Pre-processing
A custom Python script converted each DICOM to an 8-bit PNG. First, DICOM files were loaded using Python's pydicom package. We mapped pixel intensities to display values with a Value of Interest Look Up Table (VOI LUT) that used the WindowWidth and WindowCenter fields provided in the DICOM's metadata. We regulated photometric interpretation by manually setting all images to MONOCHROME2, so bones would appear as bright and the background dark. Next, we linearly rescaled the resulting pixel array to [0, 1] such that the brightest pixel had a value of 1 and the darkest a value of 0. We multiplied all values of the pixel array by 255 and cast it to an 8-bit integer, resulting in a full [0, 255] range PNG image.

### 2.3 Zero-shot Prompt
As shown in Figure 3, we developed a system prompt to guide the LVLM to correctly identify metrics of interest. The base version included a few sentences prompting the LVLM as a radiology assistant tasked with identifying one or more bones, projection view, laterality, and additional features (fractures, abnormalities) in each radiograph. Initial prompt testing was performed on the OpenAI Playground to prevent excess token usage. For development images, one out of the eleven excavation campaigns was chosen to act as a testing set, which contained images representing all classes of bones, projection views, and laterality (n = 1328). The testing images were excluded from the validation dataset.

Prompt refinement followed an error-driven process. Whenever the model produced an incorrect answer for a certain bone, projection view, or laterality, the prompt was modified in attempt to address the failure case. In ambiguous cases, where a metric in the image was unknown, an expert paleoradiologist was consulted. During prompt development, the prompt was expanded to include an explicit bone list, instructions to use the size of reference objects from the scan (e.g., paperclip), and instructions to report confidence (i.e., moderate, low if unsure). These additions led to an improvement in accuracy. Each revision was tested on a small, representative set of twenty images taken from the set of testing images.

```
You are a radiology assistant tasked with identifying bones in X-ray images from adult patients. Remember that scale is the most critical factor in identifying bones. Your job is to:
1. Identify all main bone(s) or skeletal regions in the image.
2. Determine the view (e.g., lateral, AP, axial, etc.).
3. Identify sidedness using the reference object in the image, typically a paperclip ("R" for right, "L" for left) only if necessary. If no sidedness exists, in a vertebra or cranium for example, put "N/A".
4. Note any fractures or abnormalities.
*Critical*: estimate the bone's length using the reference object as a scale reference. If the bone is not at least 3 times the length of the reference object, it is almost certainly not a large long bone.
If both sides of a limb are visible and the bone appears flat or symmetrical, the view is likely AP. If only one side is sharply outlined and the bone shows more depth or curvature, it is likely lateral. Do NOT identify a bone as a large long bone unless it is **significantly** longer than the 2 inch reference object and has clear anatomical landmarks.

Prioritize scale using the reference object over the bone shape.
Common bones to consider:
Long Bones: Metacarpals, Tibia, Femur, Humerus, Radius, Ulna, Fibula, Phalanges
Axial Bones: Lumbar vertebrae, Thoracic vertebrae, Cranium, Pelvis, Sternum, Ribs
Others: Carpal, Tarsal, Scapula, Clavicle, Mandible
Many of the large long bones, such as the femur, tibia, and humerus, look extremely similar. Before identifying these bones, think critically about the key features of the bone that the other long bones do not have. If you are unsure, report confidence as low or medium.
If multiple vertebral levels are visible (e.g., thoracic + lumbar), report as "thoracolumbar vertebrae". If uncertain about a bone's identity, report "confidence" as "low".
Output your response in this JSON format:
{
  "bone": "Metacarpals",
  "view": "AP",
  "sidedness": "Left and Right",
  "notable_features": "No obvious fractures or abnormalities",
  "confidence": "High"
}
Do not say anything else other than this. Again, make sure you consider scale using the paperclip or reference object.
```

Figure 3. Engineered zero-shot prompt queried into the system with each image. The prompt, which was iteratively revised and tested, highlights the tasks the LVLM should perform and includes key instructions that are often overlooked such as using the reference object as a scale to differentiate between bones with similar structure but different size. Names of common bones are provided to reduce ambiguity. The prompt instructs the LVLM to respond in a strict JSON format to reduce variability among responses.

### 2.4 Inference environment

We executed our prompt with the GPT-4o Vision API (June 2025 release) via the OpenAI Python SDK. We used a temperature of 0.3 to reduce variability in our responses while leaving room for creativity. We queried each image individually with the script to upload locally. Mean inference time per image was 5.2 s including network latency. The total cost was $18.92.

### 2.5 Validation protocol

We selected a random sample of 100 images (covering ten different bones) from the ten remaining sites, after excluding the testing site, for prompt development. A paleoradiologist independently reviewed the LVLM JSON outputs while simultaneously looking at the PNG files on a separate screen for validation. Three metrics were scored: (i) main-bone correct (yes/no), (ii) projection view correct, (iii) laterality correct. We computed accuracy and 95 % Wilson confidence intervals, and we assessed inter-rater agreement with Cohen's κ.

## 3. RESULTS

### 3.1 Accuracy

Out of 2219 images uploaded, processed, and received, a random sample of 100 was pulled for validation. The LVLM correctly identified the main bone in 92 of 100 images (92 %; 95 % CI 85.0–95.9 %). Projection view was correct in 80 images (80 %; 95 % CI 71.1–86.7 %), while laterality was correct in 100 images (100 %; 95 % CI 96.3.0–100.0 %). One case involving extreme fracturing and bone fragmentation was flagged as 'medium' confidence, while all others were flagged as 'high' confidence.

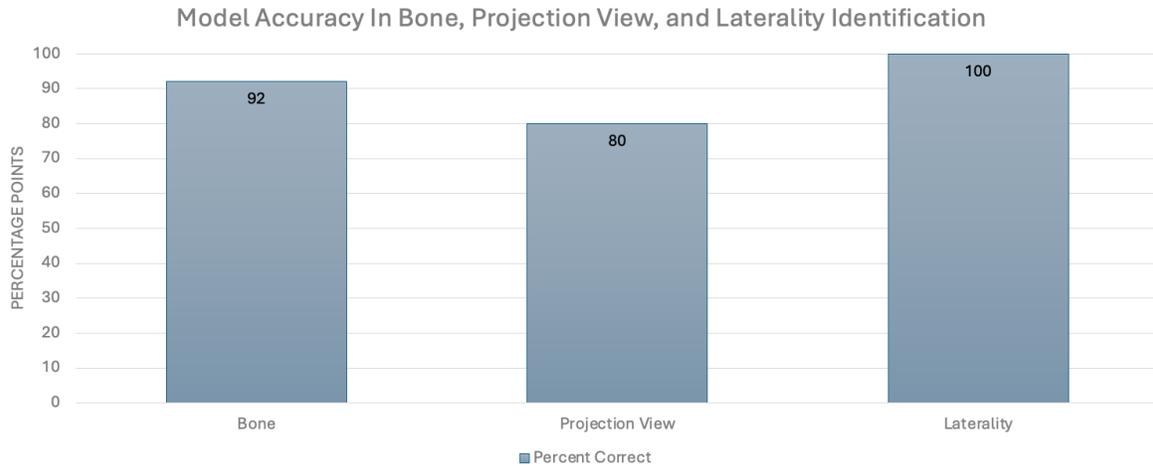

Figure 4. LVLM accuracy in bone, view, and laterality identification evaluated on a random sample of 100 images, compared to the gold standard by an expert paleoradiologist. The model achieved the highest accuracy in laterality (100% 95% CI) and lowest in projection view (80%).

**3.2 Failure modes**

8 total main-bone errors involved 4 images of misclassified metacarpals: 2 were misclassified as phalanges, 1 was misclassified as a femur, and the other misclassified as a phalanx. Furthermore, 2 images of tibiae were misclassified as femora, 1 image of thoracolumbar vertebrae was misclassified as lumbar vertebrae, and 1 image of a pelvis was misclassified as a cranium (Figures 5-6).

20 total view errors involved sixteen misclassified femoral views: 5 misclassified lateral as AP, 10 AP views misclassified as lateral, and 1 oblique view misclassified as AP. The remaining 4 errors consisted of 2 AP views of the pelvis misclassified as lateral, 1 AP view of the lumbar vertebrae misclassified as lateral, and 1 oblique view of the lumbar vertebrae misclassified as lateral (Figure 5). No hallucinations (non-skeletal objects reported as bones) were observed. To evaluate inter-rater agreement, Cohen's κ was computed for each metric. Agreement between the model and paleoradiologist was high for both bone and side identification (κ = 0.899, κ = 1.000) and moderate for view identification (κ = 0.598).

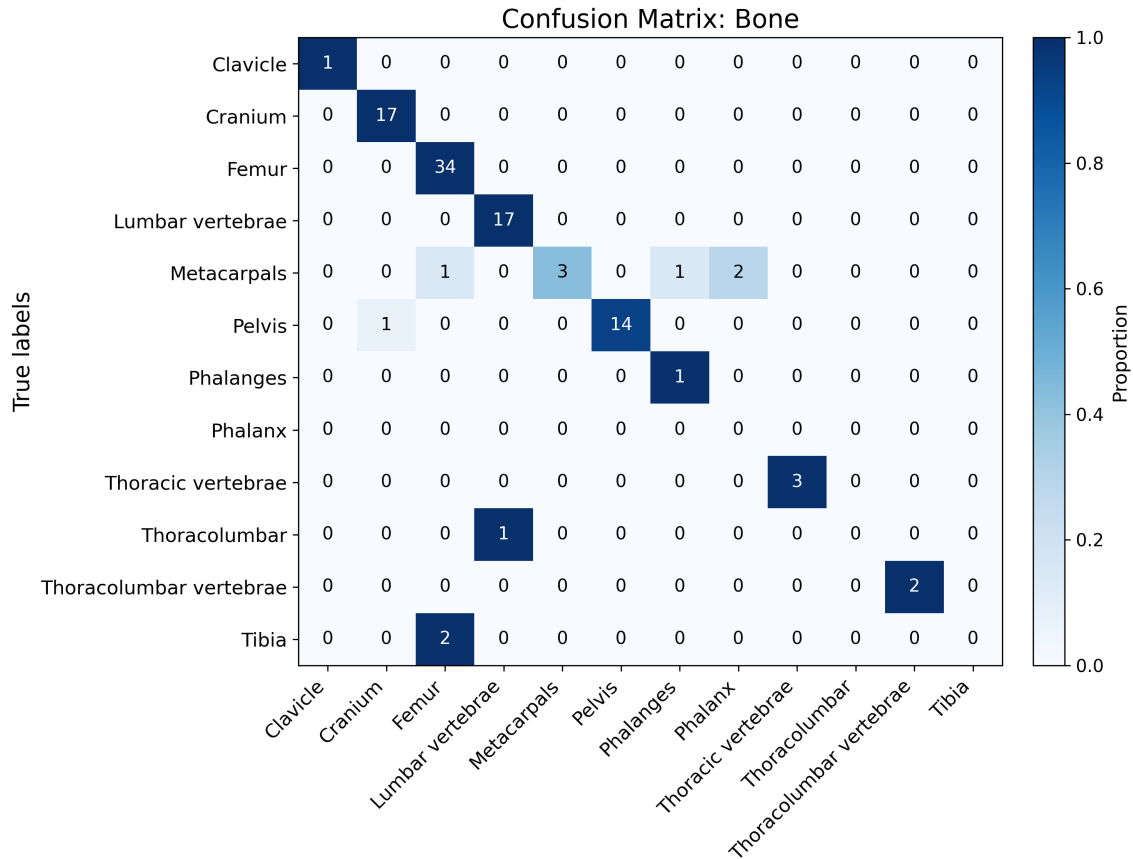

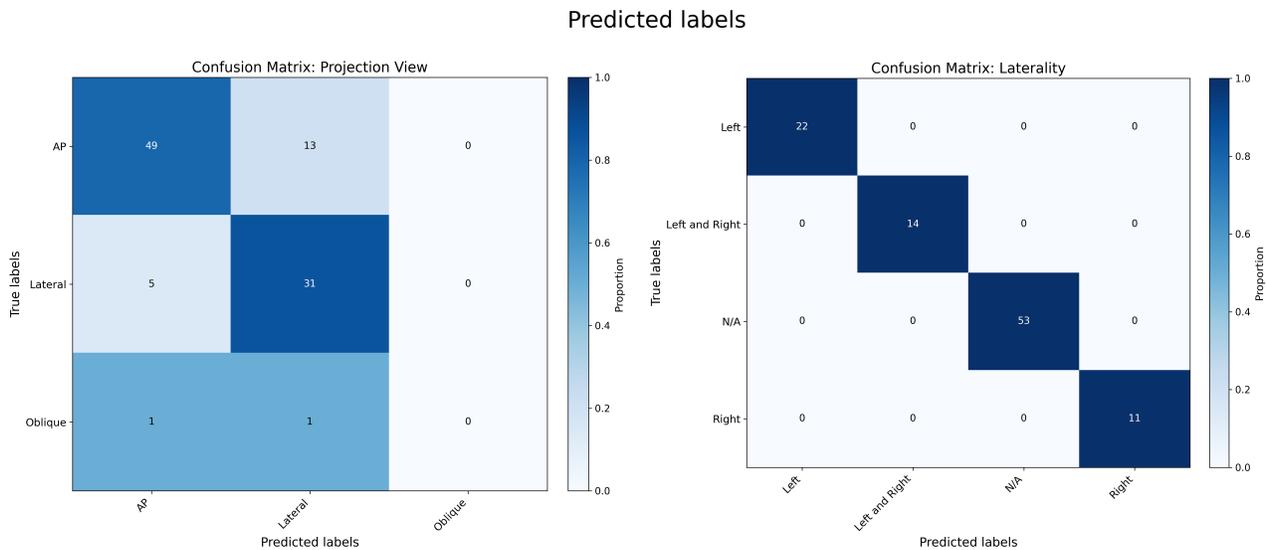

Figure 5. Confusion matrices for each measured metric (bone identification, projection view, and laterality) with the raw count of images displayed in the boxes and the shading based on the percentage correct.

### 3.3 Runtime performance.

End-to-end processing throughput averaged eleven images per minute. Horizontal scaling to four workers achieved near-linear speed-up, suggesting that the full archive could be processed overnight using standard institutional resources. Cost analysis based on current API pricing indicates an expenditure of roughly USD $0.0085 per image, for a total cost of $18.92, roughly 23-fold cheaper than a compensated expert radiologist.

## 4. DISCUSSION

Our findings demonstrate that state-of-the-art LVLMs can perform bone identification, projection view, and laterality classification on highly heterogeneous paleo-radiology images without task-specific training, achieving agreement close to that of an expert paleoradiologist. The 92 % accuracy in bone identification rivals tailored CNNs trained on modern hospital data,[9] yet our approach is unique in that it required only prompt engineering and pre-processing. By flagging ambiguous cases with low or medium confidence, the system effectively triages edge cases to human experts, enhancing accuracy and overall reliability.

The effectiveness of our approach is largely attributed to the structured, domain-specific system prompt. Ablation experiments (not shown) revealed a drop in accuracy when the explicit bone list was omitted and a further drop when instructions about the reference object in bone identification was removed. This reinforces recent literature indicating that prompt conditioning can substitute for few-shot examples in certain vision tasks.[14]

Scalability is another strength of our approach: processing the entire 8423-image archive would require roughly nine compute-hours on a single GPU node, followed by mere hours of expert review as opposed to weeks. The resulting metadata could feed directly into database interfaces, enhancing the image-based database searching in both clinical and research settings. For example, a user could upload an image of an unknown bone and the model could extract the relevant metadata to provide similar, verified images to the user. This feature would streamline research workflows by enabling more efficient searching of paleoradiology datasets.

A recurring failure case in the main bone identification seemed to be identifying metacarpal bones. More specifically, as depicted in Figure 6, in three out of the four cases, the metacarpal bone was misclassified as a phalanx of phalange, bones with very similar structure and size to a metacarpal. A recurring failure case in view identification appeared to be differentiating between an anteroposterior (AP) and lateral femoral view. This is likely due to the lack of clarification in the system prompt on how to distinguish between the two views. The prompt focuses heavily on structure and using the reference object as a scale for bone identification, whereas the main difference between femoral views is depth perception as shown in Figure 7.

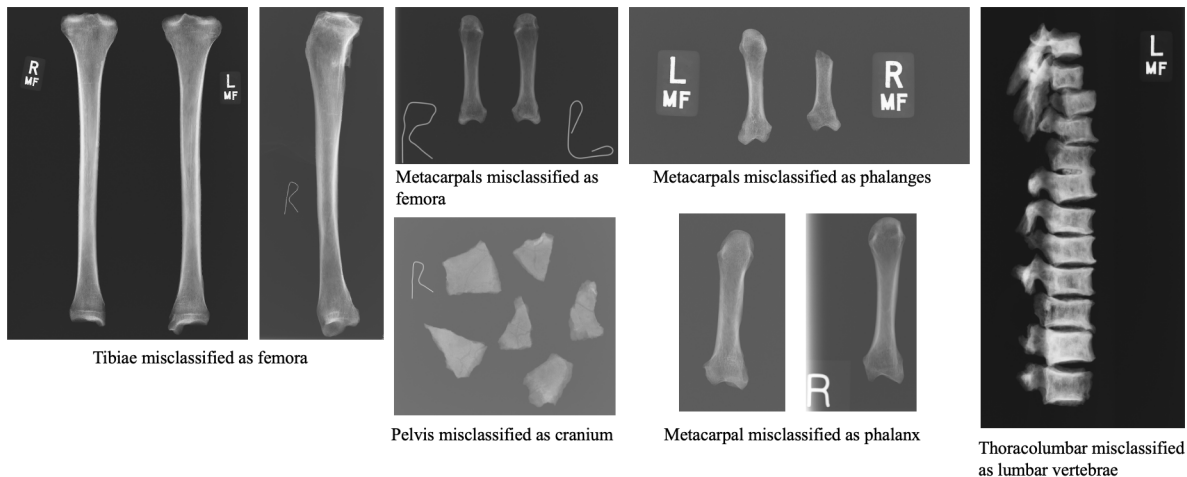

Figure 6. All eight main bone classification errors. Under each image are both the true label and predicted label. The most common error came from bones with similar structure, size, or both. Further clarification in the system prompt is needed for confident differentiation of similar bones. Other errors include an extremely fragmented bone as well as a miscounting of vertebral bodies.

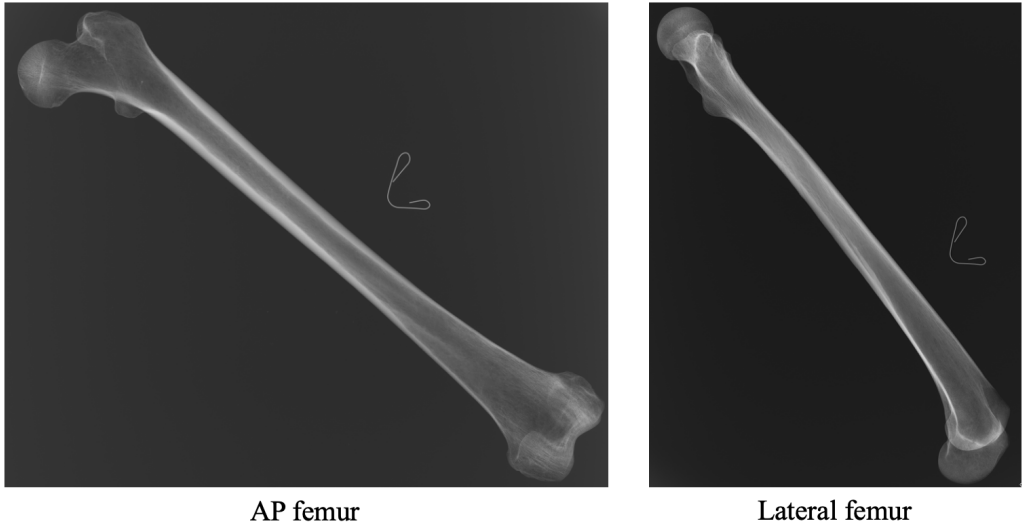

Figure 7. Anteroposterior and lateral view of the femur. In both views, the bone has a similar curvature and cortical density in the shaft. The differentiating feature is the depth at the proximal and distal ends of the femur

Beyond easing curation burdens, rapid bone identification has scholarly implications. It enables researchers to sort images into cohorts spanning epochs, ecologies, and cultural practices, thereby contextualizing skeletal pathologies within specific fields such as dietary shifts, migration patterns, and technological innovations.[10,11] For instance, an automatically generated index of femoral neck angles could be cross-referenced with isotopic evidence of activity levels to test hypotheses about labor specialization in early urban centers. Similar linkages between spinal degeneration and the rise of animal domestication have remained speculative, mainly owing to small sample sizes; high-throughput annotation can change that landscape.[15] More broadly, democratizing access to curated imaging data promotes community-engaged heritage science, allowing descendant communities to query ancestral remains without physical intrusion. These interdisciplinary opportunities underscore the urgency of scalable metadata solutions and motivate our present work.

Limitations include the modest validation sample size and dependence on cloud APIs, which may present cost or privacy barriers. Future work will include expanding the test set, exploring open-source LVLMs for on-premise deployment, and integrating pathological analysis, such as lesion identification and classification, into the system prompt.

## 5. CONCLUSION

Zero-shot prompting of a large vision–language model offers a practical, accurate, and scalable solution for curating paleo-radiology X-ray archives. By converting days of expert labeling into minutes of review, this technique opens the door to rich, extensive archaeological specimens that can further our understanding of human health and disease in the modern day. Continued refinement of prompts, confidence handling, and ontology alignment will further unlock the research potential of these invaluable historical datasets.

**This work has not been submitted for presentation or publication elsewhere.**

## ACKNOWLEDGEMENTS

This work was funded by the Vanderbilt VALIANT initiative and extended funding from Dr. Katherine D. Van Schaik. We thank the field archaeologists who collected the radiographs and the radiology reviewers for their insightful feedback. We also thank Bridget Litts for assisting in our team meetings. Artificial Intelligence was used for this project.